%% file: main.tex
\documentclass[10pt,twocolumn,letterpaper]{article}

\usepackage[
    letterpaper,
    top=1in,
    bottom=1.125in,
    left=0.8125in,
    right=0.8125in,
    columnsep=0.3125in
]{geometry}

\usepackage{newtxtext}
\usepackage{newtxmath}

\usepackage{graphicx}
\usepackage{amsmath}
\usepackage{booktabs}
\usepackage{multirow}
\usepackage{array}
\usepackage{microtype}
\usepackage{caption}
\usepackage{xspace}
\usepackage{url}
\usepackage{titlesec}
\usepackage{booktabs}
\usepackage{multirow}
\usepackage{makecell}
\usepackage{graphicx}
\usepackage[table]{xcolor}
\definecolor{headgreen}{HTML}{93C47D}
\definecolor{simclrblue}{HTML}{D9EAF7}
\definecolor{oursgreen}{HTML}{B6D7A8}
\definecolor{dinoblue}{HTML}{D9EAF7}
\definecolor{byolyellow}{HTML}{FFF2CC}
\definecolor{ourspink}{HTML}{EAD1DC}
\definecolor{barlowgreen}{RGB}{210,235,210} 
\definecolor{dinovpurple}{RGB}{225,215,245}
\definecolor{oursrow}{RGB}{235,245,235}

\usepackage[
    colorlinks=true,
    linkcolor=black,
    citecolor=black,
    urlcolor=black
]{hyperref}

\usepackage[
    capitalize,
    nameinlink
]{cleveref}

\graphicspath{{figures/}}

\titleformat{\section}
    {\normalfont\fontsize{12}{14}\selectfont\bfseries}
    {\thesection.}
    {0.5em}
    {}

\titleformat{\subsection}
    {\normalfont\fontsize{11}{13}\selectfont\bfseries}
    {\thesubsection.}
    {0.5em}
    {}

\titleformat{\subsubsection}
    {\normalfont\fontsize{10}{12}\selectfont\bfseries}
    {\thesubsubsection.}
    {0.5em}
    {}

\titlespacing*{\section}
    {0pt}{12pt}{6pt}

\titlespacing*{\subsection}
    {0pt}{10pt}{5pt}

\titlespacing*{\subsubsection}
    {0pt}{8pt}{4pt}

\begin{document}

\twocolumn[
\begin{@twocolumnfalse}

\begin{center}

    {\fontsize{16}{19}\selectfont\bfseries
    Positive Pair Geometry Matters: Optimal Transport for \\Contrastive Learning of Visual Representations
    \par}

    \vspace{10pt}

    {\fontsize{12}{14}\selectfont
    Akshit Nanda$^{1}$
    \qquad
    Shahzad Ahmad$^{2}$
    \qquad
    Ram Prasad Padhy$^{1}$
    \par}

    \vspace{6pt}

    {\fontsize{10}{12}\selectfont
    $^{1}$Department of Computer Science and Engineering, Indian Institute of Technology Bhubaneswar\\
    $^{2}$Østfold University of Applied Sciences, NTNU
    \par}

    \vspace{4pt}

    {\fontsize{9}{11}\selectfont
    \texttt{a24cs09006@iitbbs.ac.in}
    \qquad
    \texttt{shahzadnitphd@gmail.com}
    \qquad
    \texttt{ramprasad@iitbbs.ac.in}
    \par}

\end{center}

\vspace{10pt}

\noindent
{\fontsize{10}{12}\selectfont
\input{sections/abstract}
\par}

\vspace{14pt}

\end{@twocolumnfalse}
]

\input{sections/introduction}

\input{sections/relatedwork}

\input{sections/method}

\input{sections/experiments}

\input{sections/discussion}

\input{sections/conclusion}

\bibliographystyle{plain}
\bibliography{main}

\clearpage
\input{sections/supplementary}
\end{document}

%% file: sections/abstract.tex
\begin{abstract}
Contrastive self-supervised learning has achieved strong performance by learning representations from multiple augmented views of the same image. However, most existing methods construct positive pairs using independently sampled stochastic augmentations, which may alter semantic content and ignore the intrinsic geometry of the data distribution. In this work, we propose \textit{OTCLR}, an optimal transport-aware framework for contrastive learning representations that generates geometry-consistent positive samples. Instead of directly contrasting two randomly augmented views, we construct intermediate views between the original image and its augmented variants through entropic optimal-transport displacement interpolation. These transport-interpolated samples serve as positive views that better preserve image structure while explicitly modeling spatial distributional geometry. To further promote smooth representation learning, we evaluate auxiliary Sinkhorn regularization terms that encourage transport-interpolated views to remain consistent with their endpoint images. The proposed method can be incorporated into standard contrastive learning pipelines without modifying the encoder architecture. Experiments on multiple benchmark datasets show that our approach improves representation quality and transfer learning performance compared with conventional augmentation-based contrastive learning baselines.
\end{abstract}

%% file: sections/introduction.tex
\section{Introduction}
\label{sec:intro}
Self-supervised visual representation learning has emerged as a powerful paradigm for learning semantic features from unlabeled data \cite{jing2020self,lecun2022path}. Among existing approaches, contrastive learning methods have demonstrated that strong visual representations can be learned by maximizing agreement between different augmented views of the same image while contrasting them against other samples in the batch \cite{oord2018representation,he2020momentum,chen2020simple}. The success of these methods has made data augmentation a central design component in modern self-supervised 
\begin{figure}
    \centering
    \includegraphics[width=\linewidth]{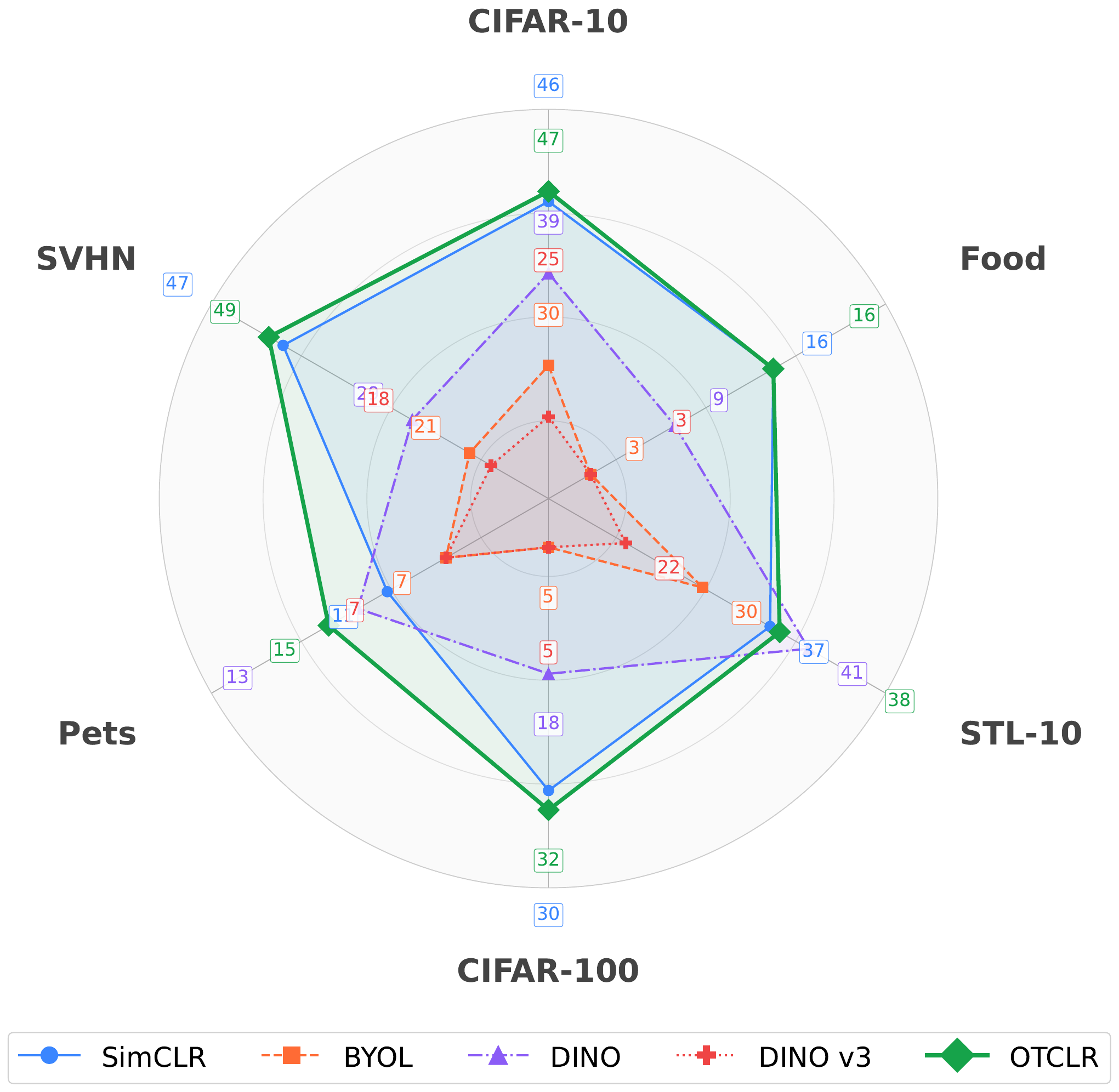}
    \caption{Linear evaluation comparison across diverse downstream tasks. We report the top-1 accuracy (\%) using a linear probe on six benchmark datasets: CIFAR-10, Food, STL-10, CIFAR-100, Pets, and SVHN. Compared to prominent self-supervised baselines including SimCLR, BYOL, DINO, and DINO v3.}
    \label{fig:trailer}
\end{figure}
learning pipelines \cite{chen2020simple,tian2020makes}.

Despite their effectiveness, existing contrastive frameworks rely heavily on stochastic augmentations to construct positive pairs \cite{chen2020simple,he2020momentum,grill2020bootstrap}. In these methods, independently sampled augmentations are assumed to preserve the same semantic identity and therefore provide a meaningful learning signal \cite{chen2020simple,tian2020makes}. However, this assumption introduces an important trade-off. Weak augmentations often produce overly similar views that lead to trivial alignment, whereas aggressive augmentations can generate visually abrupt transformations that distort semantic structure and complicate optimization \cite{tian2020makes,purushwalkam2020demystifying,wang2021understanding}. Consequently, current methods largely ignore the geometry of the transition between positive views, even though the encoder is encouraged to learn invariance across these transformations.

In this work, we investigate whether positive-pair generation can be made more geometry-aware using optimal transport (OT) \cite{villani2009optimal}. Instead of directly contrasting independently sampled augmentations, we generate intermediate views by computing entropic Wasserstein barycenters between the original view and strongly augmented views \cite{cuturi2013sinkhorn,cuturi2014fast}. The resulting samples lie along an approximate Wasserstein geodesic between image distributions and provide transport-smoothed intermediate representations that preserve smoother semantic transitions between views \cite{benamou2015iterative,solomon2015convolutional}.

The proposed formulation can be interpreted as modifying the geometry of view generation rather than changing the underlying encoder architecture or contrastive objective. Our approach is fully compatible with standard contrastive learning frameworks \cite{chen2020simple,he2020momentum} and introduces no additional inference-time architectural complexity, since all transport computations are restricted to the pretraining stage. 

Linear evaluations on various benchmarking datasets (Figure~\ref{fig:trailer}) demonstrate that geometry-aware view generation improves representation quality compared to conventional augmentation-based contrastive learning. Our analysis further suggests that incorporating transport-based geometric structure into view construction offers a promising approach to improving self-supervised visual representation learning. \textbf{We claim the following contributions:}
\begin{itemize}
    \item We propose a geometry-aware positive-view generation strategy for
    contrastive learning based on optimal-transport displacement interpolation
    between the source image and the strongly augmented image views.


    \item We incorporate an image-space Sinkhorn regularization term that
    constrains each generated view to remain close to both its weak anchor and
    corresponding strong augmentation endpoint.

    \item We demonstrate through transfer-learning, clustering, and
    visualization experiments that the proposed approach improves representation
    quality across multiple downstream benchmarks, while preserving the
    architecture and inference-time computational footprint of the base encoder.
\end{itemize}

%% file: sections/relatedwork.tex
\section{Related Work}
\label{sec:related work}
\textbf{Contrastive self-supervision.}
Contrastive learning has emerged as one of the most successful paradigms for self-supervised visual representation learning \cite{jing2020self,lecun2022path}. Early approaches such as Contrastive Predictive Coding (CPC) demonstrated that discriminative objectives can learn meaningful representations from unlabeled data \cite{oord2018representation}. Subsequently, SimCLR showed that a simple framework based on strong data augmentation, large batch sizes, and a nonlinear projection head can achieve competitive performance without architectural modifications \cite{chen2020simple}. Similar developments include MoCo, which improves negative-sample utilization through a momentum encoder and memory queue \cite{he2020momentum}. These methods share a common principle: representations are learned by maximizing agreement between positive views while contrasting them against negative samples \cite{oord2018representation,chen2020simple,he2020momentum}. In contrast to modifying the contrastive objective itself, our work focuses on the construction of positive views and investigates whether geometry-aware view generation can provide a richer learning signal.\\
\noindent
\textbf{Non-contrastive and clustering-based SSL.}
Recent self-supervised approaches have demonstrated that strong visual representations can be learned without explicit negative samples. BYOL employs a bootstrapped prediction framework using online and target networks \cite{grill2020bootstrap}, while methods such as SwAV leverage online clustering objectives to learn semantically meaningful representations \cite{caron2020unsupervised}. Although these approaches differ substantially from contrastive methods in their learning objectives, they continue to rely heavily on the quality and diversity of augmented views \cite{chen2020simple,grill2020bootstrap,caron2020unsupervised}. Consequently, improvements in view generation are largely orthogonal to the choice of self-supervised objective and may benefit both contrastive and non-contrastive frameworks.\\
\noindent
\textbf{Optimal transport in vision.}
Optimal transport (OT) provides a principled framework for comparing probability distributions by accounting for the cost of transporting mass between them \cite{villani2009optimal}. Entropic regularization and Sinkhorn-based solvers have made OT practical for large-scale machine learning applications \cite{cuturi2013sinkhorn,benamou2015iterative} and have led to widespread adoption in computer vision, domain adaptation, generative modeling, and representation learning. Unlike pixel-wise similarity measures, Wasserstein distances capture spatial displacement and geometric structure within images \cite{villani2009optimal}. This property makes OT particularly attractive for modeling transformations that involve spatial movement, such as cropping, translation, and local appearance changes. Our work leverages this geometric perspective to construct transport-based intermediate views for self-supervised learning.\\
\noindent
\textbf{Interpolation-Based View Generation.}
Data augmentation is central to modern self-supervised learning, where semantic identity is expected to remain invariant under transformations such as cropping, color distortion, blurring, and horizontal flipping \cite{chen2020simple,tian2020makes}. Existing augmentation pipelines generate positive pairs through independent stochastic transformations and do not explicitly model the relationship between the resulting views \cite{chen2020simple,he2020momentum}. Interpolation-based approaches instead seek meaningful intermediate samples that bridge multiple augmentations. While linear interpolation in pixel space provides a simple mechanism 
\begin{figure*}[h!]
    \centering
    \includegraphics[width=0.7\linewidth]{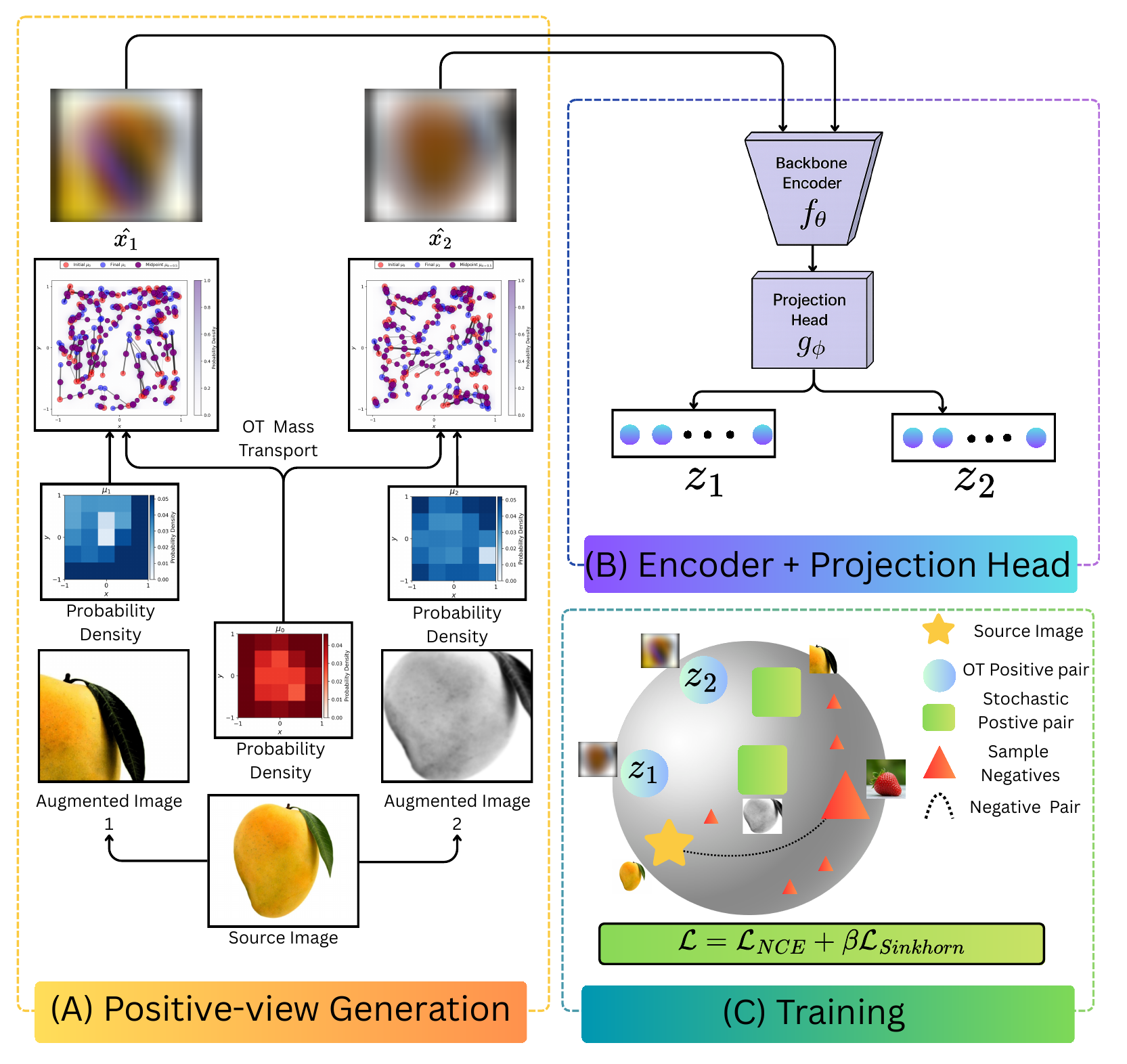}
    \caption{
Overview of the proposed OTCLR approach for self-supervised learning.
(A) Positive-view generation: a source view and two strongly augmented views are interpreted as image-space probability distributions. Entropic optimal transport is used to move mass from the source view toward each strong augmentation, producing two OT-displacement positive views.
(B) Encoder and projection head: the generated positive views are passed through a shared backbone encoder $f_\theta$ and projection head $g_\phi$, followed by normalization to obtain contrastive embeddings $z_1$ and $z_2$.
(C) Training objective: the embeddings are optimized with the bidirectional NT-Xent contrastive loss, where the paired OT-displacement views form positives and other batch samples act as negatives. An image space Sinkhorn regularization term $\mathcal{L}_{Sinkhorn}$ is added to the contrastive objective, giving $\mathcal{L}=\mathcal{L}_{\mathrm{NCE}}+\beta\mathcal{L}_{Sinkhorn}$.
}
    \label{fig:method}
\end{figure*}
for generating such samples, it fails to account for spatial correspondence and object displacement. In contrast, Wasserstein barycentric interpolation constructs intermediate views by explicitly modeling the transport of image mass, yielding geometry-aware samples that better respect the underlying structure of visual transformations \cite{cuturi2013sinkhorn}.\\
\noindent
\textbf{Position of this work.}
The proposed approach lies at the intersection of self-supervised learning and optimal transport. Unlike prior work that primarily modifies network architectures or learning objectives, we focus on the geometry of positive-view generation. Specifically, we replace independently sampled positive views with transport-smoothed intermediate views obtained through Wasserstein barycentric interpolation \cite{cuturi2014fast}. This design allows the encoder architecture, optimization procedure, and contrastive objective to remain unchanged, enabling a controlled study of how geometry-aware view construction influences representation learning and downstream transfer performance.

%% file: sections/method.tex
\section{Methodology}
\label{sec:methodology}

\subsection{Overview}
We propose an image-space optimal-transport (OT) \cite{peyre2019computational} view generator for
view-based self-supervised learning (SSL). The method modifies only the
positive-view construction stage and is compatible with any contrastive SSL framework
that learns from multiple augmented views of the same image. All
downstream components—encoder, projection head, and optimizer remain unchanged. We instantiate the method within SimCLR \cite{chen2020simple}, but
the view generator is framework agnostic and can be applied to other
view-based contrastive SSL pipelines.

Standard SimCLR \cite{chen2020simple} independently samples two strong augmented views of each
image and treats them as a positive pair. Our method constructs two strong augmented views,
\begin{equation*}
\scriptsize
    x_{s,k} = t_{s,k}(x), \quad k \in \{1,2\},
    \label{eq:view_sampling}
\end{equation*}
where $t_{s,k}$ is the standard SimCLR
strong augmentation pipeline applied on the source image $x$. For each source–strong pair $(x, x_{s,k})$,
we solve an entropic OT problem on a low-resolution image grid and
generate an intermediate displacement view via bilinear splatting \cite{khamlich2025optimal}. The
two resulting views $\hat{x}_1$ and $\hat{x}_2$ replace the standard
positive pair in the InfoNCE \cite{oord2018representation} objective. The OT module is used only at
training time and introduces no additional parameters or inference
overhead.

\subsection{View-Based SSL Formulation}

Let $F_\Theta$ denote a view-based SSL comprising an encoder $f_\theta$
and a projection head $g_\phi$ \cite{chen2020simple}. View-based SSL methods optimize
\begin{equation*}
\scriptsize
    \mathcal{L}_{\mathrm{SSL}}
    =
    \mathbb{E}_{x}
    \bigl[
    \ell_{\mathrm{SSL}}
    \bigl(F_\Theta(v_1), \ldots, F_\Theta(v_K)\bigr)
    \bigr],
    \qquad
    v_k = t_k(x).
    \label{eq:generic_ssl}
\end{equation*}
where each view $v_k$ is produced by a stochastic augmentation
$t_k \sim \mathcal{T}$, where $\mathcal{T}$ is a family of stochastic augmentations. Our contribution lies exclusively in the construction of $v_k$; the model $F_\Theta$ is unmodified.

In the SimCLR instantiation, a shared encoder $f_\theta$ and projection
head $g_\phi$ produce $\ell_2$-normalized embeddings:
\begin{equation}
\scriptsize
    z(v) =
    \frac{g_\phi(f_\theta(v))}
    {\bigl\|g_\phi(f_\theta(v))\bigr\|_2}.
    \label{eq:normalized_projection}
\end{equation}
Given a minibatch of $B$ images and two views per image, the InfoNCE
loss \cite{oord2018representation} is
\begin{equation}
\scriptsize
    \mathcal{L}_{\mathrm{NCE}}
    =
    \frac{1}{2B}
    \sum_{i=1}^{2B}
    -\log
    \frac{
    \exp\!\bigl(z_i^\top z_{p(i)} / \tau\bigr)
    }{
    \displaystyle\sum_{j \neq i}
    \exp\!\bigl(z_i^\top z_j / \tau\bigr)
    },
    \label{eq:infonce}
\end{equation}
where $p(i)$ is the positive index paired with sample $i$ and $\tau$ is
the temperature.

\subsection{Image-Space Optimal Transport}

We represent each image view as a discrete probability measure on a
bounded spatial grid, following the standard discrete OT formulation \cite{peyre2019computational}. Given a view $u \in [0,1]^{H \times W \times C}$,
each channel is resized to an $m \times m$ grid
$\Omega_m = \{r_i\}_{i=1}^{N}$ with $N = m^2$ and $r_i \in [0,1]^2$ denotes the normalized two dimensionl spatial coordinate of the $i$-th grid location. The resized channel is then shifted to be non-negative and normalized to unit mass, yielding a probability vector $p \in \Delta^{N-1}$:
\begin{equation}
\scriptsize
    \mu_p = \sum_{i=1}^{N} p_i \,\delta_{r_i}, \qquad
    \Delta^{N-1} =
    \Bigl\{p \in \mathbb{R}_{+}^{N} : p^\top \mathbf{1} = 1\Bigr\}.
    \label{eq:image_measure}
\end{equation}
where, $\mu_p$, $\delta_{r_i}$, and $\Delta$ are the probability measure, Dirac delta at location $r_i$, and $(N-1)$-dimensional probability simplex respectively.
The pairwise transport cost $\mathbf{(C_{ij})}$ between grid locations is the normalized
squared Euclidean distance:
\begin{equation}
\scriptsize
    \mathbf{C_{ij}} =
    \frac{\|r_i - r_j\|_2^2}{\max_{a,b}\|r_a - r_b\|_2^2}.
    \label{eq:ground_cost}
\end{equation}

For the source view and strong-view, let $p^c$ and
$p_{s,k}^c$ denote the probability vectors obtained from their $c$-th channels. We solve the entropy-regularized OT problem using sinkhorn iterations \cite{cuturi2013sinkhorn,peyre2019computational}:
\begin{equation}
\scriptsize
    \Pi_{k,c}^{\star}
    =
    \mathop{\arg\min}_{\Pi \in U(p^c,\,p_{s,k}^c)}
    \left[
    \langle \Pi, \mathbf{C} \rangle
    -
    \varepsilon H(\Pi)
    \right]
    \label{eq:entropic_ot}
\end{equation}
where $\Pi_{k,c}^{\star}$ is the entropy-regularized optimal transport plan for the $c$-th channel of the $k$-th source-augmentation pair. $H(\Pi)=-\sum_{i,j}\Pi_{ij}(\log\Pi_{ij}-1)$ is the negative
entropy, and the transport polytope is
\begin{equation}
\scriptsize
    U(p^c, p_{s,k}^c) =
    \bigl\{
    \Pi \in \mathbb{R}_{+}^{N \times N} :
    \Pi\mathbf{1} = p^c,\;
    \Pi^\top\mathbf{1} = p_{s,k}^c
    \bigr\}.
    \label{eq:transport_polytope}
\end{equation}
The entropy regularization enables efficient batched computation via
Sinkhorn iterations on low-resolution grids.

\subsection{OT Displacement View Generation}

The optimal plan $\Pi_{k,c}^{\star}$ encodes how pixel mass is transported from
the source view $x$ to the strong view $x_{s,k}$. Rather than using
either endpoint directly, we generate an intermediate view by placing
each transported mass packet at an interpolated spatial location, following the displacement-interpolation principle used for synthetic
state generation in reduced-order modeling \cite{khamlich2025optimal}. For
interpolation parameter $\alpha_k \in [0,1]$, mass transported from
grid point $r_i$ to $r_j$ is deposited at
\begin{equation*}
\scriptsize
    r_{ij}(\alpha_k) = (1-\alpha_k)r_i + \alpha_k r_j.
    \label{eq:displacement_location}
\end{equation*}

The channel-wise displaced measure is:
\begin{equation}
\scriptsize
    \nu_{\alpha_k,\varepsilon}^{c}
    =
    \sum_{i=1}^{N}\sum_{j=1}^{N}
    (\Pi_{k,c}^{\star})_{ij}\,
    \delta_{r_{ij}(\alpha_k)}.
    \label{eq:displaced_measure}
\end{equation}

Since $r_{ij}(\alpha_k)$ need not coincide with a grid vertex, mass is
redistributed to neighboring vertices via bilinear splatting, a forward-warping operation related to differentiable splatting and image
warping \cite{niklaus2020softmax}. Letting
$B_\ell(r)$ denote the bilinear basis weight from location $r$ to vertex
$r_\ell$, the discretized channel grid is:
\begin{equation}
\scriptsize
    \bar{p}_{\alpha_k,\ell}^{c}
    =
    \sum_{i=1}^{N}\sum_{j=1}^{N}
    (\Pi_{k,c}^{\star})_{ij}\,
    B_\ell\!\bigl(r_{ij}(\alpha_k)\bigr),
    \qquad \ell = 1,\ldots,N.
    \label{eq:bilinear_splatting}
\end{equation}
Here, $\bar{p}_{\alpha_k,\ell}^{c}$ is the reconstructed mass at grid vertex $r_{\ell}$ for channel $c$ of the generated view.
The generated view is formed by stacking the reconstructed channels,
upsampling to the encoder input resolution, and clipping to the valid
image range:
\begin{equation}
\scriptsize
    \hat{x}_k = D_{\alpha_k,\varepsilon}(x, x_{s,k}),
    \qquad k \in \{1,2\}.
    \label{eq:generated_views}
\end{equation}
where D denotes the composition of bilinear splatting, channel stacking,
upsampling, and clipping.
The anchor or source view promotes structural consistency, while the strong
endpoint preserves the semantic invariances induced by standard
augmentations. Together, the two generated views form a
geometry-aware positive pair as shown in Figure~\ref{fig:method}.

\subsection{Training Objective}
The proposed objective replaces the two standard augmented
views in the Eq.~\ref{eq:infonce} with the generated OT displacement views $\hat{x}_{1}$ and $\hat{x}_{2}$ from Eq.~\ref{eq:generated_views}:
\begin{equation}
\scriptsize
    \mathcal{L}_{\mathrm{OTD}}
    =
    \mathcal{L}_{\mathrm{NCE}}(\hat{x}_1, \hat{x}_2).
    \label{eq:otd_objective}
\end{equation}
where $\mathcal{L}_{\mathrm{NCE}}(\hat{x}_1,\hat{x}_2)$ denotes Eq.~\ref{eq:infonce}
evaluated after substituting $\hat{x}_1,\hat{x}_2$ for the batch of standard
augmented views, prior to the embedding step in
Eq.~\ref{eq:normalized_projection}.
We also consider a Sinkhorn-regularized variant that discourages the
generated views from drifting away from either the anchor or the
corresponding strong endpoint. Let
$S_\varepsilon(\cdot,\cdot)$ denote the entropic Sinkhorn OT cost \cite{peyre2019computational,cuturi2013sinkhorn},
computed channel-wise on the same low-resolution image grid by mapping
each image argument to its probability vector as in
Eq.~\ref{eq:image_measure}. The auxiliary regularizer is
\begin{equation}
\scriptsize
    S_\varepsilon(a,b)
=
\min_{\gamma \in U(p_a^c,p_b^c)}
\sum_{i,j}
\gamma_{ij} \mathbf{C}_{ij}
+
\varepsilon
\sum_{i,j}
\gamma_{ij}(\log \gamma_{ij}-1),
\end{equation}
where $a,b\in[0,1]^{H\times W\times C}$ be two images, and let $p_a^c$ and $p_b^c$ denote the probability vectors obtained from their $c$-th channels using Eq.~\ref{eq:image_measure}.
\begin{equation}
\scriptsize
    \mathcal{L}_{\mathrm{Sinkhorn}}
    =
    \frac{1}{2}
    \Big[\sum_{k} S_\varepsilon(\hat{x}_k, x) + \sum_{k} S_\varepsilon(\hat{x}_k, x_{s,k})\Big].
\end{equation}
\noindent
The final pretraining objective is

\begin{equation}
\scriptsize
    \mathcal{L}_{\mathrm{pre}}
    =
    \mathcal{L}_{\mathrm{OTD}}(\hat{x}_1,\hat{x}_2)
    +
    \beta \mathcal{L}_{\mathrm{Sinkhorn}}.
    \label{eq:regularized_objective}
\end{equation}
where $\beta$ controls the influence of the Sinkhorn regularization.


%% file: sections/experiments.tex
\section{Experiments}
\label{sec:experiments}

\subsection{Training Details}
\textbf{Datasets:} We evaluate our method \textit{OTCLR} on different publicly available benchmarking datasets: CIFAR-10, CIFAR-100 ~\cite{krizhevsky2009learning}, Street View House Numbers (SVHN) ~\cite{netzer2011reading}, STL-10 ~\cite{coates2011analysis}, Food-101 ~\cite{bossard2014food}, and Oxford-IIIT Pets ~\cite{parkhi2012cats}. These datasets differ in visual domain, object granularity, scene complexity, and category structure, covering natural objects, fine-grained categories, digit recognition, texture-rich food images, and low-resolution object-centric scenes.\\
\textbf{Experimental Setup:} We implement our framework in TensorFlow ~\cite{abadi2016tensorflow} and pre-train a ResNet-18 encoder ~\cite{he2016deep} on ImageNet-300, a 1000-class subset of ImageNet ~\cite{russakovsky2015imagenet} containing approximately 300,000 images. We train all models for 20 epochs with SGD and momentum, and evaluate at batch sizes of 256 and 512. Representation quality is assessed via linear evaluation across multiple downstream benchmarks, reporting top-1 and top-5 classification accuracies.\\
\textbf{Implementation Details:}
The encoder backbone is ResNet-18 ~\cite{he2016deep}, followed by a two-layer projection head as in SimCLR \cite{chen2020simple}. During pretraining, images are augmented using random resized cropping, horizontal flipping, color jittering, and Gaussian blur. The NT-Xent loss is optimized using SGD with momentum 0.9 and weight decay $10^{-4}$. The temperature parameter is set to $\tau=0.5$. For OT interpolation, the OT is computed on a $16 \times 16$ grid with regularization $\epsilon = 0.05$, $T = 20$ sinkhorn iterations, and the interpolation coefficient is fixed at $\alpha=0.5$, corresponding to the midpoint between the source image and its augmented counterpart \cite{cuturi2014fast}. All experiments are conducted on a single NVIDIA L40 GPU. \\
\begin{figure*}[t!]
    \centering
    \includegraphics[width=\linewidth]{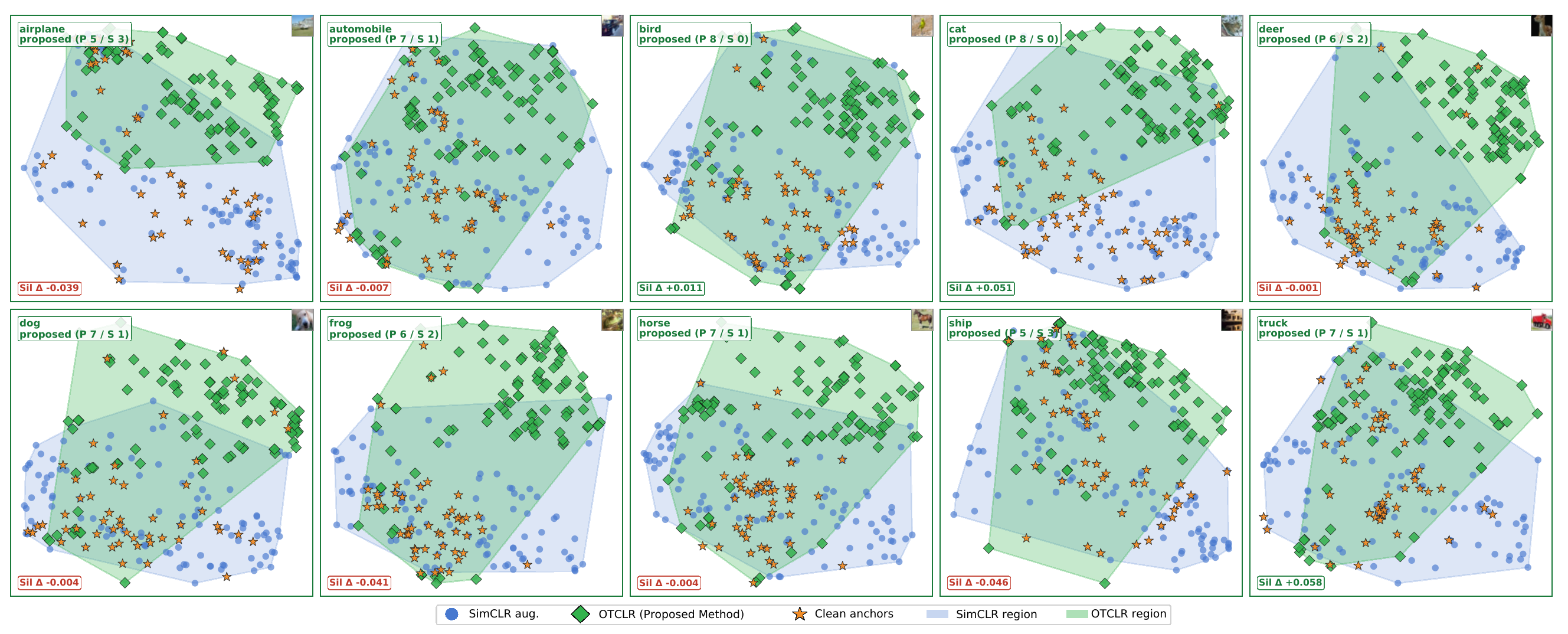}

\caption{\textbf{Class-wise t-SNE visualization of CIFAR-10 augmentation representations.} Blue circles, green diamonds, and orange stars denote SimCLR views, OTCLR views, and clean anchor samples, respectively; shaded regions indicate their feature-space coverage. Compared with SimCLR, OTCLR produces more structured and balanced class-wise coverage and generates representations that are more robust to augmentation variability and potentially more generalizable. The top-left inset reports the proposed method (P) versus SimCLR (S) for the metrics in Table~\ref{tab:cifar10_aug_quality}; the top-right inset shows a representative image for each class, and the bottom-left inset reports the class-wise change in silhouette score.}
    \label{fig:tsne}
\end{figure*}

\begin{table*}[t!]
\centering
\setlength{\tabcolsep}{3pt}
\resizebox{\textwidth}{!}{
\input{tables/augmentation_quality}
}
\caption{CIFAR-10 augmentation quality comparison using 500 samples. Best values are in bold.}
\label{tab:cifar10_aug_quality}
\end{table*}
\noindent
\textbf{Model Complexity:}
The proposed method OTCLR operates exclusively in the view-generation stage and does not modify the backbone architecture, projection head, or downstream classifier. Consequently, the number of trainable parameters and the inference-time computational complexity remain unchanged. For the MoCo instantiation considered in this work \cite{he2020momentum}, both vanilla MoCo and the proposed OT Displacement variant employ 23.01M parameters during pretraining and 11.182M parameters during linear evaluation, with identical computational costs of 0.154 GFLOPs per image pair and 0.074 GFLOPs per image, respectively. The additional computational overhead is confined to the generation of transport-based views during pretraining. As shown in Table~\ref{tab:complexity}, OT Displacement increases the average training step time from 21.50 ms to 30.87 ms and reduces throughput from 11,908 to 8,292 image pairs/s, while maintaining a comparable GPU memory footprint. Since transport computations are performed only during training, model size, memory requirements, and computational complexity during linear evaluation and inference remain unchanged.

\begin{table}[htbp!]
    \centering
    \scriptsize
    \setlength{\tabcolsep}{2pt}

    \resizebox{\columnwidth}{!}{
    \input{tables/complexity}
    }
    \caption{Computational complexity comparison}
    \label{tab:complexity}
\end{table}

\begin{table*}[htbp!]
    \centering
    \scriptsize
    \setlength{\tabcolsep}{3pt}
    \renewcommand{\arraystretch}{0.4}

    \resizebox{0.9\textwidth}{!}{%

\input{tables/full_linear_eval}
    }
    \caption{The proposed method was compared with multiple SSL models and the linear evaluation was compared on various benchmarking datasets and the top-1 and top-5 results are reported.}
    \label{tab:model_compare}
\end{table*}

\noindent
\textbf{Evaluation Metrics:} Downstream performance is measured using top-1 and top-5 classification accuracy over the complete held-out test split. We also report representation-level clustering quality to evaluate whether the learned embeddings form semantically meaningful groups without using class labels during feature extraction. For this, we compute the Silhouette score \cite{rousseeuw1987silhouettes}, which quantifies intra-cluster
compactness relative to inter-cluster separation; Compactness
measures the average dispersion of embeddings around their cluster centroid;
Centroid Margin measures the separation between the nearest neighbors
cluster centroids, and the Inter-Intra cluster distance ratio
captures the overall separation between semantic groups relative to within-group
compactness, following standard internal
cluster validity criteria based on compactness and separation \cite{dunn1974well,davies1979cluster,calinski1974dendrite}. To characterize the behavior of the OT view generator
specifically, we further report the View Distance (View Dist.) and the View
Cosine similarity (View Cos.), measuring embedding-space distance and angular alignment
between the two generated views $\hat{x}_1$ and $\hat{x}_2$;
Anchor Distance (Anchor Dist.), measuring the embedding space distance between each
generated view and the source image $x$; and pixel $L_2$, measuring
raw pixel-space deviation between the generated views and their corresponding
strong-augmentation endpoints $x_{s,k}$. 
The detailed definitions and formulas for
all the metrics are provided in the \textcolor{blue}{\textit{supplementary section}}.

\subsection{Results and Ablation}

\textbf{Representation Quality Analysis}:\\
Table~\ref{tab:cifar10_aug_quality} evaluates the quality of augmentation representations produced by SimCLR and the proposed OT Displacement method on CIFAR-10. The proposed method OTCLR yields more compact and consistent views, reducing class compactness from 0.0633 to 0.0544 and view-pair feature distance from 71.78 to 34.79. Correspondingly, view-pair cosine similarity improves from 0.9958 to 0.9982, showing that two augmented views of the same image remain closely aligned in the learned representation space.

The OT-based views also reduce the anchor feature distance from 59.60 to 56.69 and pixel-level augmentation distance from 0.34 to 0.17, indicating improved semantic preservation with less excessive visual distortion. Although SimCLR achieves a marginally higher global silhouette score, both scores are close to zero and negative, making this metric less informative in the present setting. Figure~\ref{fig:tsne} provides a class-wise t-SNE visualization that qualitatively supports these results: OT-displacement views form more concentrated neighborhoods around their corresponding clean anchors than standard SimCLR views. Overall, the results demonstrate that transport-based view generation produces more stable, compact, and semantically consistent augmentations.

\noindent \textbf{Linear Classification Performance:}\\
Table~\ref{tab:model_compare} compares the proposed OT Displacement approach with representative self-supervised learning methods OTCLR on multiple downstream benchmarks. The proposed method achieves the best performance on CIFAR-10, CIFAR-100, Food-101, Oxford-IIIT Pets, STL-10, and SVHN, outperforming SimCLR \cite{chen2020simple}, DINO \cite{caron2021emerging}, DINOv3 \cite{simeoni2025dinov3}, and BYOL \cite{grill2020bootstrap} across both Top-1 and Top-5 accuracy metrics. These results suggest that incorporating transport-aware geometric information during view generation leads to more transferable visual representations across diverse recognition tasks.

\begin{figure*}[htbp!]
    \centering
    \includegraphics[width=\linewidth]{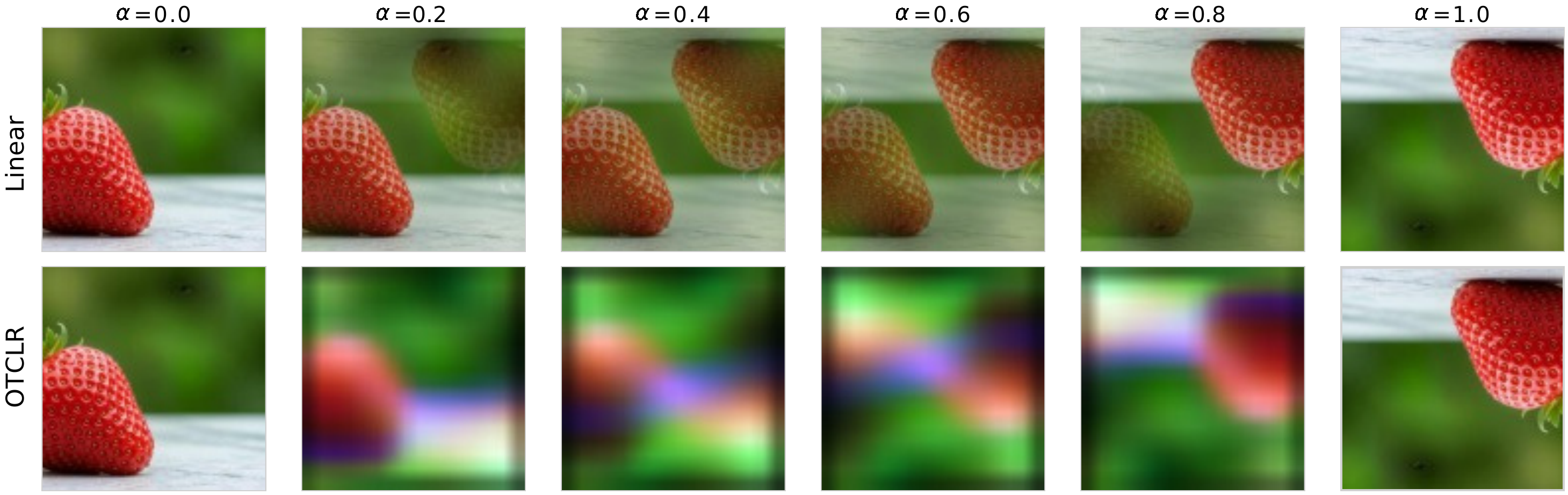}
    \caption{
Comparison of linear interpolation and the proposed OTCLR
interpolation on a strawberry image across interpolation coefficients
$\alpha$. Linear interpolation blends pixel intensities directly, producing
ghosted intermediate images, whereas OTCLR transports image mass
along the spatial transport path and yields more coherent intermediate views.
}
    \label{fig:interpolation}
\vspace{-2mm}
\end{figure*}

\noindent \textbf{Qualitative Analysis:}\\
Figure~\ref{fig:interpolation} visualizes the intermediate views generated by the proposed transport-based interpolation strategy. Unlike pixel-wise interpolation, which produces unrealistic blurry transitions and progressively loses structural information, Wasserstein-based interpolation preserves object boundaries and spatial correspondences throughout the interpolation path. OT displacement interpolation maintains the geometric structure of the image while producing smooth semantic transitions between the source and target views. These visualizations illustrate how transport-based view generation can produce meaningful intermediate samples that better preserve image geometry and object structure than vanilla linear interpolation.

\begin{figure}[htbp!]
    \centering
    \includegraphics[width=\linewidth]{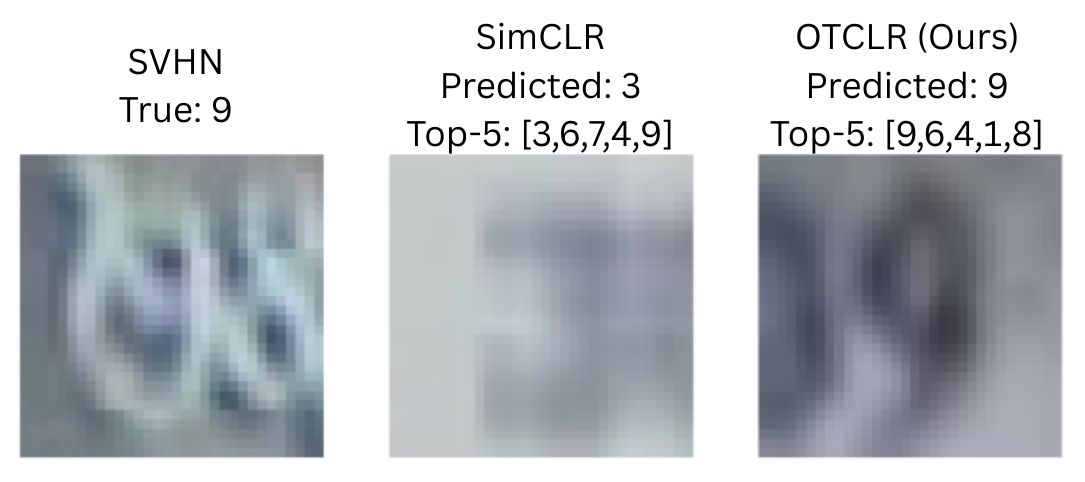}
    \caption{Qualitative comparison of predictions produced by SimCLR and the proposed OT-based model on a challenging SVHN sample. The input image contains the digit "9", which the proposed method correctly identifies, whereas SimCLR predicts the digit "3". The top-5 predictions are also shown for each model. The result suggests that transport-based view generation can improve feature robustness under ambiguous visual conditions and lead to more discriminative representations.}
    \label{fig:prediction}
\end{figure}

To gain further insight into the learned representations, Figure~\ref{fig:prediction} presents a qualitative comparison between SimCLR \cite{chen2020simple} and the proposed OT-based approach on a challenging SVHN \cite{netzer2011reading} example. Although the image exhibits substantial blur and low visual quality, the proposed method correctly predicts the target digit, whereas SimCLR misclassifies it. Furthermore, the correct label appears at a higher rank among the top-5 predictions produced by the proposed model. This observation suggests that transport-based view generation encourages the encoder to learn features that are more robust to distortions and variations in appearance. While qualitative in nature, these examples complement the quantitative improvements observed in the linear evaluation experiments.
\begin{table*}[t]
    \centering
    \scriptsize
    \setlength{\tabcolsep}{3pt}
    \renewcommand{\arraystretch}{0.8}

    \resizebox{\textwidth}{!}{%

\input{tables/half_linear_eval}
    }
    \caption{We compared our method with the baseline SimCLR for different batch sizes and pre-trained on different datasets by computing the top-1 and top-5 scores from linear evaluation on varied benchmarking datasets.}
    \label{tab:linear_eval_batch}
    \vspace{2mm}
\end{table*}

\noindent \textbf{Ablation Studies:}\\
Table~\ref{tab:linear_eval_batch} reports linear evaluation results under different pretraining datasets and batch sizes. Across all experimental settings, the proposed method consistently improves downstream performance relative to the corresponding baseline. The improvements are observed for both CIFAR-10 and ImageNet-300 pretraining and remain stable across batch sizes of 256 and 512. Notably, the gains are particularly pronounced on CIFAR-100 and Food-101, indicating that transport-based view generation is especially beneficial for challenging datasets with higher intra-class variability. The consistent improvements across datasets and training configurations demonstrate that the effectiveness of the proposed approach is not tied to a specific pretraining distribution or optimization setting.

%% file: tables/augmentation_quality.tex
\begin{tabular}{lcccccccc}
\toprule
\textbf{Method} &
\textbf{Silhouette $\uparrow$} &
\textbf{Compactness $\downarrow$} &
\textbf{Centroid Margin $\uparrow$} &
\textbf{Inter/Intra $\uparrow$} &
\textbf{View Dist. $\downarrow$} &
\textbf{View Cos. $\uparrow$} &
\textbf{Anchor Dist. $\downarrow$} &
\textbf{Pixel L2 $\downarrow$} \\
\midrule
SimCLR \cite{chen2020simple}&
\textbf{-0.0440} &
0.0633 &
-0.0044 &
0.9302 &
71.7879 &
0.9958 &
59.6021 &
0.3413 \\
\cellcolor{oursrow}OTCLR (Ours) &
\cellcolor{oursrow}-0.0461 &
\cellcolor{oursrow}\textbf{0.0544} &
\cellcolor{oursrow}\textbf{-0.0038} &
\cellcolor{oursrow}\textbf{0.9307} &
\cellcolor{oursrow}\textbf{34.7904} &
\cellcolor{oursrow}\textbf{0.9982} &
\cellcolor{oursrow}\textbf{56.6968} &
\cellcolor{oursrow}\textbf{0.1791} \\
\bottomrule
\end{tabular}

%% file: tables/complexity.tex
\begin{tabular}{llcccc}
\toprule
\textbf{Stage} & \textbf{Method} & \textbf{Params (M)} & \textbf{GFLOPs} & \textbf{Step (ms)} & \textbf{Thrpt.} \\
\midrule

\multirow{2}{*}{Pretraining}
& MoCo \cite{he2020momentum}& 23.01 & 0.154 & 21.50 & 11,908 \\
& \cellcolor{oursrow}OTCLR (Ours) & \cellcolor{oursrow}23.01 & \cellcolor{oursrow}0.154 & \cellcolor{oursrow}30.87 & \cellcolor{oursrow}8,292 \\

\midrule

\multirow{2}{*}{Linear Eval.}
& MoCo \cite{he2020momentum} & 11.18 & 0.074 & 4.48 & 57,143 \\
& \cellcolor{oursrow}OTCLR(Ours) & \cellcolor{oursrow}11.18 & \cellcolor{oursrow}0.074 & \cellcolor{oursrow}6.00 & \cellcolor{oursrow}42,678 \\

\bottomrule
\end{tabular}

%% file: tables/full_linear_eval.tex
\begin{tabular}{lcccccccccccc}
\toprule
\multirow{2}{*}{\textbf{Method}} &
\multicolumn{2}{c}{\makecell{\textbf{CIFAR-10}}} &
\multicolumn{2}{c}{\makecell{\textbf{CIFAR-100}}} &
\multicolumn{2}{c}{\makecell{\textbf{Food-101}}} &
\multicolumn{2}{c}{\makecell{\textbf{Pets}}} &
\multicolumn{2}{c}{\makecell{\textbf{STL-10}}} &
\multicolumn{2}{c}{\makecell{\textbf{SVHN}}} \\
\cmidrule(lr){2-3}
\cmidrule(lr){4-5}
\cmidrule(lr){6-7}
\cmidrule(lr){8-9}
\cmidrule(lr){10-11}
\cmidrule(lr){12-13}
& \textbf{Top-1} & \textbf{Top-5}
& \textbf{Top-1} & \textbf{Top-5}
& \textbf{Top-1} & \textbf{Top-5}
& \textbf{Top-1} & \textbf{Top-5}
& \textbf{Top-1} & \textbf{Top-5}
& \textbf{Top-1} & \textbf{Top-5} \\
\midrule

SimCLR \cite{chen2020simple}
& 46.09 & \textbf{91.40}
& 30.47 & 50.00
& \textbf{16.40} & \textbf{31.25}
& 10.93 & 32.03
& 36.71 & 81.25
& 46.87 & 76.56 \\

DINO \cite{caron2021emerging}
& 39.30 & 87.42
& 18.14 & 42.28
& 9.41 & 25.84
& 12.74 & \textbf{38.58}
& \textbf{41.38} & \textbf{90.30}
& 28.59 & 75.04 \\

BYOL \cite{grill2020bootstrap}
& 30.28 & 81.35
& 5.27 & 19.25
& 3.45 & 13.54
& 6.61 & 23.38
& 30.25 & 83.39
& 20.63 & 63.68 \\

DINOv3 \cite{simeoni2025dinov3}
& 25.36 & 75.92
& 4.60 & 16.59
& 3.36 & 12.59
& 7.23 & 27.80
& 21.86 & 72.84
& 17.80 & 62.49 \\

\midrule
\rowcolor{oursrow}
\textbf{OTCLR(Ours)}
& \textbf{46.87} & 90.62
& \textbf{31.25} & \textbf{52.34}
& \textbf{16.40} & 30.46
& \textbf{14.84} & 35.15
& 37.50 & 70.90
& \textbf{48.43} & \textbf{79.60} \\

\bottomrule
\end{tabular}

%% file: tables/half_linear_eval.tex
\begin{tabular}{lllcccccc}
\toprule
\multirow{2}{4pt}{\textbf{Pretraining Dataset}} &
\multirow{2}{*}{\textbf{Batch Size}} &
\multirow{2}{*}{\textbf{Method}} &
\multicolumn{6}{c}{\textbf{Linear Evaluation}} \\
\cmidrule(lr){4-9}
& & &
\makecell{\textbf{CIFAR-10}\\\textbf{Top-1 / Top-5}} &
\makecell{\textbf{CIFAR-100}\\\textbf{Top-1 / Top-5}} &
\makecell{\textbf{Food-101}\\\textbf{Top-1 / Top-5}} &
\makecell{\textbf{Pets}\\\textbf{Top-1 / Top-5}} &
\makecell{\textbf{STL-10}\\\textbf{Top-1 / Top-5}} &
\makecell{\textbf{SVHN}\\\textbf{Top-1 / Top-5}} \\
\midrule

\multirow{4}{*}{CIFAR-10}
& \multirow{2}{*}{512}
& SimCLR \cite{chen2020simple}
& 45.31 / 91.40
& 21.09 / 48.43
& 8.59 / 23.43
& \textbf{10.15} / 28.12
& 34.37 / \textbf{82.81}
& 38.28 / 78.90 \\

& 
& \cellcolor{oursrow}\textbf{OTCLR(Ours)}
& \cellcolor{oursrow}\textbf{46.87} / \textbf{92.18}
& \cellcolor{oursrow}\textbf{30.46} / \textbf{50.00}
& \cellcolor{oursrow}\textbf{17.18} / \textbf{32.03}
& \cellcolor{oursrow}\textbf{10.15} / \textbf{35.93}
& \cellcolor{oursrow}\textbf{35.03} / 81.25
& \cellcolor{oursrow}\textbf{48.43} / \textbf{79.68} \\

& \multirow{2}{*}{256}
& SimCLR \cite{chen2020simple}
& 42.96 / 86.71
& 20.31 / 49.21
& 10.90 / 25.78
& 10.93 / 28.12
& 33.59 / 83.59
& 43.70 / \textbf{81.25} \\

&
& \cellcolor{oursrow}\textbf{OTCLR(Ours)}
& \cellcolor{oursrow}\textbf{46.09} / \textbf{92.96}
& \cellcolor{oursrow}\textbf{28.12} / \textbf{50.78}
& \cellcolor{oursrow}\textbf{16.40} / \textbf{31.25}
& \cellcolor{oursrow}\textbf{13.28} / \textbf{33.59}
& \cellcolor{oursrow}\textbf{35.15} / \textbf{85.15}
& \cellcolor{oursrow}\textbf{47.65} / 78.12 \\

\midrule

\multirow{4}{*}{ImageNet300}
& \multirow{2}{*}{512}
& SimCLR \cite{chen2020simple}
& 46.09 / \textbf{91.40}
& 30.47 / 50.00
& \textbf{16.40} / \textbf{31.25}
& 10.93 / 32.03
& \textbf{36.71} / 81.25
& \textbf{46.87} / 76.56 \\

&

& \cellcolor{oursrow}\textbf{OTCLR(Ours)}
& \cellcolor{oursrow}\textbf{46.87} / 90.62
& \cellcolor{oursrow}\textbf{31.25} / \textbf{52.34}
& \cellcolor{oursrow}15.62 / \textbf{31.25}
& \cellcolor{oursrow}\textbf{13.28} / \textbf{38.28}
& \cellcolor{oursrow}35.93 / \textbf{82.03}
& \cellcolor{oursrow}\textbf{46.87} / \textbf{78.12} \\

& \multirow{2}{*}{256}
& SimCLR \cite{chen2020simple}
& 47.65 / 89.06
& 22.65 / 41.40
& 12.50 / 31.25
& 11.71 / 31.20
& 33.59 / 80.46
& 36.71 / 80.46 \\

&
& \cellcolor{oursrow}\textbf{OTCLR(Ours)}
& \cellcolor{oursrow}\textbf{47.65} / \textbf{91.40}
& \cellcolor{oursrow}\textbf{29.68} / \textbf{50.78}
& \cellcolor{oursrow}\textbf{13.28} / \textbf{32.81}
& \cellcolor{oursrow}\textbf{16.40} / \textbf{34.37}
& \cellcolor{oursrow}\textbf{35.15} / \textbf{83.59}
& \cellcolor{oursrow}\textbf{45.31} / \textbf{81.25} \\
\bottomrule
\end{tabular}

%% file: sections/discussion.tex
\section{Discussion}
\label{sec:discussion}

The experimental results show that geometry-aware positive-view generation improves self-supervised representations without changing the encoder architecture or contrastive learning framework. Unlike standard augmentations that sample two views independently, the proposed OT-displacement strategy constructs views along an image-space transport path between the anchor and strong augmentations, preserving semantic content while introducing meaningful appearance variation. The clustering metrics and class-wise t-SNE visualizations indicate that the proposed views produce more compact and better separated representations. This supports the hypothesis that positive-pair geometry plays an important role in contrastive representation learning. The Sinkhorn regularizer further stabilizes training by keeping each generated view close to both the anchor and its corresponding strong augmentation, preventing unrealistic drift from the transport path. Since the OT module is used only during pretraining, the method introduces no additional parameters or inference-time cost. Its main limitation is the extra training-time computation required for Sinkhorn iterations and displacement-view generation.

%% file: sections/conclusion.tex
\section{Conclusion}
\label{sec:conclusion}
We presented an optimal-transport displacement framework for improving
positive view construction in contrastive self-supervised learning. Instead
of relying solely on independently sampled stochastic augmentations, the
proposed method constructs intermediate image-space views along an
OT-induced displacement path between the anchor and the strong augmented
views. These generated views are then used directly in the standard InfoNCE
objective, with an optional Sinkhorn regularization term that stabilizes the
transport-based view generation.

Across the evaluated datasets, the proposed method produces representations
with improved cluster compactness, stronger semantic separation, and more
coherent class-wise t-SNE structure compared with the baseline. These
results indicate that transport-aware positive pairs can provide a more
structured training signal for contrastive learning. Importantly, the method
does not modify the encoder or add inference-time overhead, making it a
drop-in training-time enhancement for view-based self-supervised learning.

%% file: sections/supplementary.tex
\section*{Supplementary Material}

\subsection*{Evaluation Metrics}
We evaluate our method OTCLR using three categories of metrics: (i) classification performance, (ii) representation quality, and (iii) OT view generation quality. This section provides the definitions and mathematical formulations of all evaluation metrics used throughout the paper.

Classification performance is evaluated using Top-1 and Top-5 accuracies on the held-out test split. Representation quality is assessed using the Silhouette Score \cite{rousseeuw1987silhouettes}, Compactness, Centroid Margin, and the Inter-Intra Cluster Distance Ratio \cite{calinski1974dendrite,davies1979cluster,dunn1974well}. To characterize the behavior of the proposed OT view generator, we additionally report View Distance, View Cosine Similarity, Anchor Distance, and Pixel $L_2$.

Let $\{z_i\}_{i=1}^{n}$ denote the $\ell_2$-normalized embeddings of the evaluation set, partitioned into $K$ clusters $\{\mathcal{C}_1,\ldots,\mathcal{C}_K\}$ obtained using $k$-means clustering, with corresponding cluster centroids $\{c_1,\ldots,c_K\}$. 

For the OT view generation metrics, let $x$ denote the original input
image, $\hat{x}_1$ and $\hat{x}_2$ denote the two generated OT
displacement views corresponding to $x$, and $z(\cdot)$ denote the
corresponding $\ell_2$-normalized embedding.

\subsection*{Classification Metrics}

\paragraph{Top-1 Accuracy.}
Top-1 accuracy measures the percentage of samples for which the class
with the highest predicted probability matches the ground-truth label:
\[
\hat y_i=\arg\max_c p_i(c),
\]
\[
\mathrm{Top1}
=
\frac{1}{n}
\sum_{i=1}^{n}
\mathbf{1}
\left(\hat y_i=y_i
\right),
\]
where $p_i(c)$ denotes the predicted probability for class $c$,
$y_i$ is the ground-truth label, and
$\mathbf{1}(\cdot)$ is the indicator function.

\paragraph{Top-5 Accuracy.}
Top-5 accuracy measures the percentage of samples for which the
ground-truth label appears among the five highest-scoring predicted
classes:
\[
\mathrm{Top5}
=
\frac{1}{n}
\sum_{i=1}^{n}
\mathbf{1}
\left(
y_i \in \mathrm{Top5}(p_i)
\right),
\]
where $\mathrm{Top5}(p_i)$ denotes the set of five classes with the
largest predicted probabilities.

\subsection*{Representation Quality Metrics}

\paragraph{Silhouette Score~\cite{rousseeuw1987silhouettes}.}
For an embedding $z_i$ in cluster $\mathcal{C}_k$, let $a(z_i)$ be the
mean distance to all other points in $\mathcal{C}_k$, and $b(z_i)$ be the
mean distance to all points in the nearest other cluster. The silhouette
coefficient is
\[
  s(z_i) = \frac{b(z_i) - a(z_i)}{\max\{a(z_i),\,b(z_i)\}},
\]
and the reported score is the mean $\bar{s} = \frac{1}{n}\sum_i s(z_i)$,
with $\bar{s} \in [-1, 1]$; higher values indicate better-separated, more
compact clusters.
 
\paragraph{Compactness~\cite{davies1979cluster}.}
For each cluster $\mathcal{C}_k$ with centroid
$c_k = \frac{1}{|\mathcal{C}_k|}\sum_{z_i \in \mathcal{C}_k} z_i$,
compactness is the average squared distance of cluster members to their
centroid:
\[
  \mathrm{Compactness} = \frac{1}{K}\sum_{k=1}^{K}
  \frac{1}{|\mathcal{C}_k|}\sum_{z_i \in \mathcal{C}_k} \|z_i - c_k\|_2^2.
\]
Lower values indicate tighter, more concentrated clusters.
 
\paragraph{Centroid Margin~\cite{davies1979cluster}.}
Centroid margin measures the minimum Euclidean separation between each cluster centroid and its nearest neighboring centroid. For each centroid $c_k$, we compute the distance to its nearest
neighboring centroid:
\[
  \mathrm{Margin}(c_k) = \min_{j \neq k} \|c_k - c_j\|_2,
\]
and report the mean margin across all clusters,
$\overline{\mathrm{Margin}} = \frac{1}{K}\sum_{k=1}^{K}
\mathrm{Margin}(c_k)$. Higher values indicate better separated cluster
centers.
 
\paragraph{Inter-Intra Cluster Distance Ratio~\cite{dunn1974well}.}
Let the intra-cluster distance be the mean pairwise distance within
clusters,
\[
  D_{\mathrm{intra}} = \frac{1}{K}\sum_{k=1}^{K}
  \frac{2}{|\mathcal{C}_k|(|\mathcal{C}_k|-1)}
  \sum_{\substack{z_i,\,z_j \in \mathcal{C}_k \\ i < j}}
  \|z_i - z_j\|_2
\]
and let the inter-cluster distance be the mean pairwise distance between
centroids,
\[
  D_{\mathrm{inter}} = \frac{2}{K(K-1)}\sum_{k < l} \|c_k - c_l\|_2.
\]
The reported ratio is $R = D_{\mathrm{inter}} / D_{\mathrm{intra}}$;
higher values indicate better-separated, more distinct semantic groups
relative to their internal spread.

\subsection*{OT View Generation Metrics}
\paragraph{View Distance.}
For each image $x$ in the evaluation batch, let $z(\hat{x}_1)$ and
$z(\hat{x}_2)$ denote the $\ell_2$-normalized embeddings of its two
generated OT displacement views. View
distance is the mean Euclidean distance between paired view embeddings:
\[
  \mathrm{ViewDist} = \frac{1}{n}\sum_{i=1}^{n}
  \big\| z(\hat{x}_1^{(i)}) - z(\hat{x}_2^{(i)}) \big\|_2.
\]
Lower values indicate that the generated positive pair maps to nearby
points in embedding space.
 
\paragraph{View Cosine Similarity.}
Using the same paired view embeddings, the view cosine similarity is
\[
  \mathrm{ViewCos} = \frac{1}{n}\sum_{i=1}^{n}
  z(\hat{x}_1^{(i)})^\top z(\hat{x}_2^{(i)}),
\]
measuring angular alignment between the two generated views, with values
in $[-1, 1]$; higher values indicate stronger semantic agreement between
the positive pair.
 
\paragraph{Anchor Distance.}
Let $z(x)$ denote the embedding of the unaugmented source image $x$.
Anchor distance measures the mean embedding-space distance between each
generated view and its source:
\[
  \mathrm{AnchorDist} = \frac{1}{2n}\sum_{i=1}^{n}\sum_{k=1}^{2}
  \big\| z(\hat{x}_k^{(i)}) - z(x^{(i)}) \big\|_2.
\]
This quantifies the average displacement of the generated
views from the original image in embedding space; lower values indicate
stronger structural consistency with the source.
 
\paragraph{Pixel $L_2$.}
Unlike the preceding metrics, pixel $L_2$ is computed directly in image
space rather than embedding space. For each generated view $\hat{x}_k$ and
its corresponding strong-augmentation endpoint $x_{s,k}$,
\[
  \mathrm{PixelL2} = \frac{1}{2n}\sum_{i=1}^{n}\sum_{k=1}^{2}
  \big\| \hat{x}_k^{(i)} - x_{s,k}^{(i)} \big\|_2,
\]
where each image is flattened into a vector of pixel intensities before computing the Euclidean norm. This measures the raw visual deviation introduced by the OT displacement and bilinear splatting steps relative to the standard strong augmentation, independent of any learned representation. Lower values indicate that the generated views remain visually closer to the corresponding strong augmentation.